\documentclass[10pt,twocolumn,letterpaper]{article}

\usepackage{cvpr}
\usepackage{times}
\usepackage{epsfig}
\usepackage{graphicx}
\usepackage{amsmath}
\usepackage{amssymb}

\usepackage{stfloats}
\usepackage{cuted}
\usepackage{booktabs}
\usepackage{multirow}

\usepackage[breaklinks=true,bookmarks=false]{hyperref}

\cvprfinalcopy 

\def\cvprPaperID{9} 

\ifcvprfinal\fi
\begin{document}

\title{BatVision with GCC-PHAT Features for Better Sound to Vision Predictions}

\author{Jesper Haahr Christensen\\
Technical University of Denmark\\
{\tt\small jehchr@elekro.dtu.dk}
\and
Sascha Hornauer\\
UC Berkeley / ICSI\\
{\tt\small saschaho@icsi.berkeley.edu}
\and 
Stella Yu \\
UC Berkeley / ICSI \\
{\tt\small stellayu@berkeley.edu}
}

\maketitle

\begin{figure}[b!]
\includegraphics[width=0.99\linewidth]{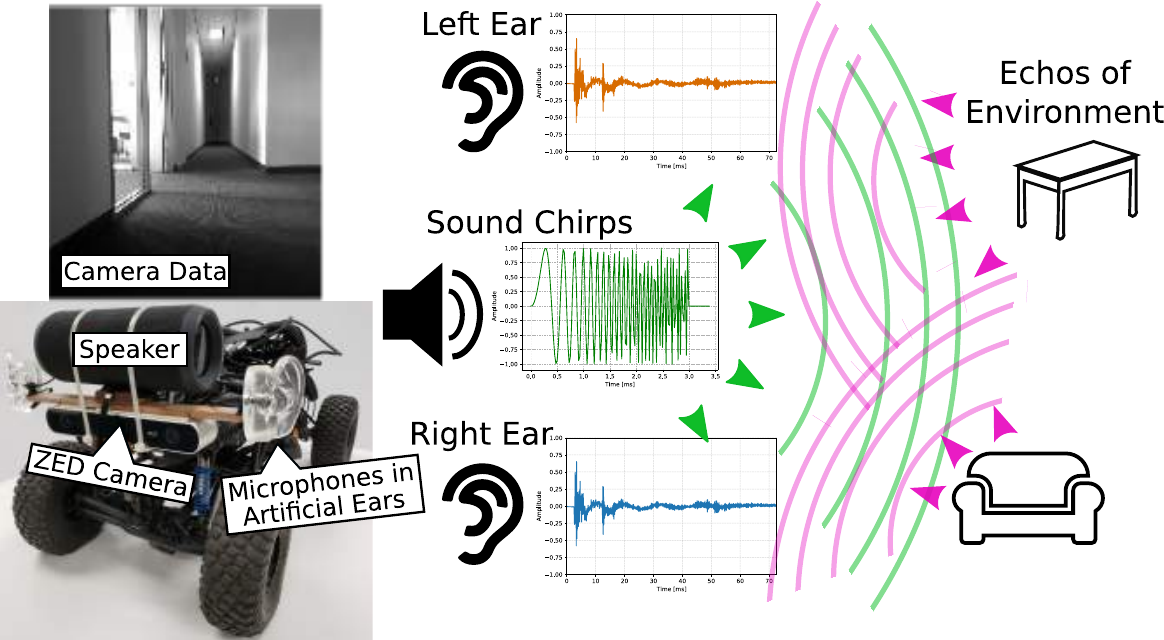}
\caption{The {\it BatVision} \cite{christensen2019batvision} system learns to generate visual scenes by just listening to echos with two ears. Mounted on a model car, the system has two microphones embedded into artificial human ears, a speaker, and a stereo camera which is {\it only used during training} for providing visual image ground-truth. The speaker emits sound chirps in an office space and the microphones receive echos returned from the environment. The camera captures stereo image pairs, based on which depth maps can be calculated.
\label{fig:overview}}
\end{figure}


\section{Introduction}

\begin{figure}[b!]
\includegraphics[width=0.95\linewidth]{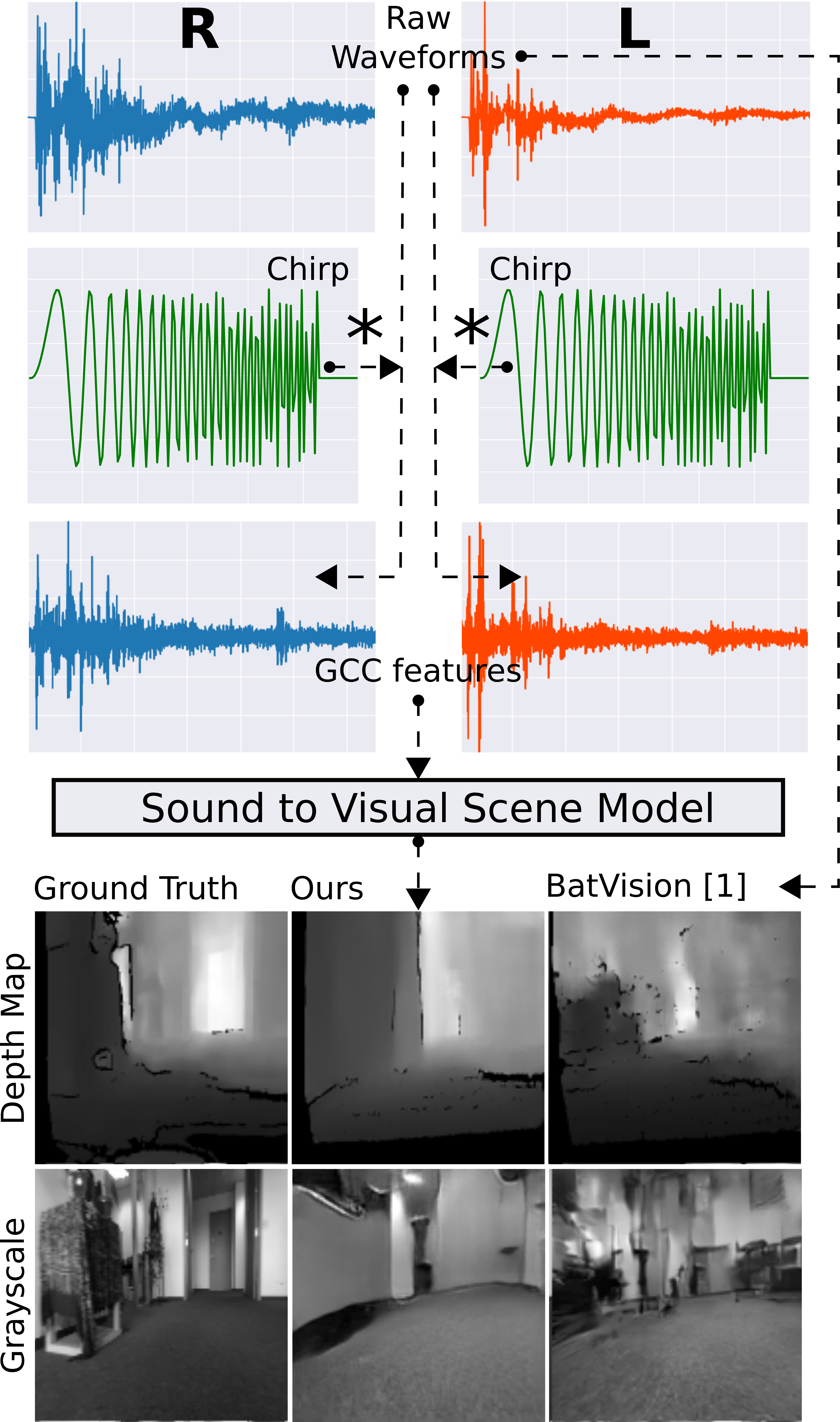}
\caption{Generalized Cross-Correlation Features (GCC) contribute largely to improved reconstruction performance. Each left and right waveform is independently correlated ({\huge \textbf{$*$}}) with the sent chirp signal (see equation \ref{eqn:gcc}). Using only these features as input in our network shows clearer depth and grayscale images with less artifacts and a more plausible room layout.
\label{fig:gcc}}
\end{figure}

We present a method for learning associations between binaural sound signals and visual scenes. Our task is to train a machine learning system that can turn binaural sound signals to 1) 3D depth maps and 2) grayscale images of plausible layout of the scene ahead.

Solving this task can benefit robot navigation and machine vision with complementary information or enable a new sensor modality in no-light conditions.

Our inspiration for this work comes from nature, where bats, dolphins and whales utilize acoustic information heavily. They adapted to environments where light is sparse. Bats have evolved advanced ears (pinnae) that provides vision in the dark known as {\it echolocation}: They sense the world by continuously emitting ultrasonic pulses and process echos returned from the environment and prey. Likewise, humans suffering from vision loss have shown to develop capabilities of echolocation using palatal clicks similar to dolphins, learning to sense obstacles in the 3D space by listening to the returning echoes ~\cite{human_echo,HumanUltrasonicEcholocation}.

Trying to harness sound for artificial systems, previous work also mimics parts of biological systems. By using an artificial pinnae pair of bats, highly reflecting ultrasonic targets in the 3D space were located. The ears act as complex direction-dependent spectral filters {\it and} head-related transfer functions have been modelled to better mimic how a particular ear (left or right) receives sound from a point in space~\cite{Biomimetic_Sonar,3DEcholocation}.

We investigate how to visualize the full 3D layout ahead only from binaural echos, recorded from microphones in artificial ears. Sound chirps are played from a speaker into the environment which we also record with a stereo camera. With the time-paired data of generated depth-images and echos, we train a network to predict the former from the latter. As a proof of concept we also predict monocular grayscale images with the objective of generating plausible layout of free space and obstacles. We show an overview of our proposed system in Fig.~\ref{fig:overview}.

Our contribution is an enhanced sound-to-visual system using generalized cross-correlation (GCC) features which we compare to raw waveforms and spectrograms as input encoding (cmp. Fig. \ref{fig:gcc}).
We further show the advantage of Residual-in-Residual Dense Blocks ~\cite{DBLP:journals/corr/abs-1809-00219} for the generator in our architecture. We also introduce spectral normalization~\cite{DBLP:journals/corr/abs-1802-05957} to the PatchGAN~\cite{pix2pix} discriminator to replace batch normalization and empirically observe a more stabilized training process.

\section{Audio-Visual Dataset}
\begin{figure}
    \centering
    \includegraphics[width=\linewidth]{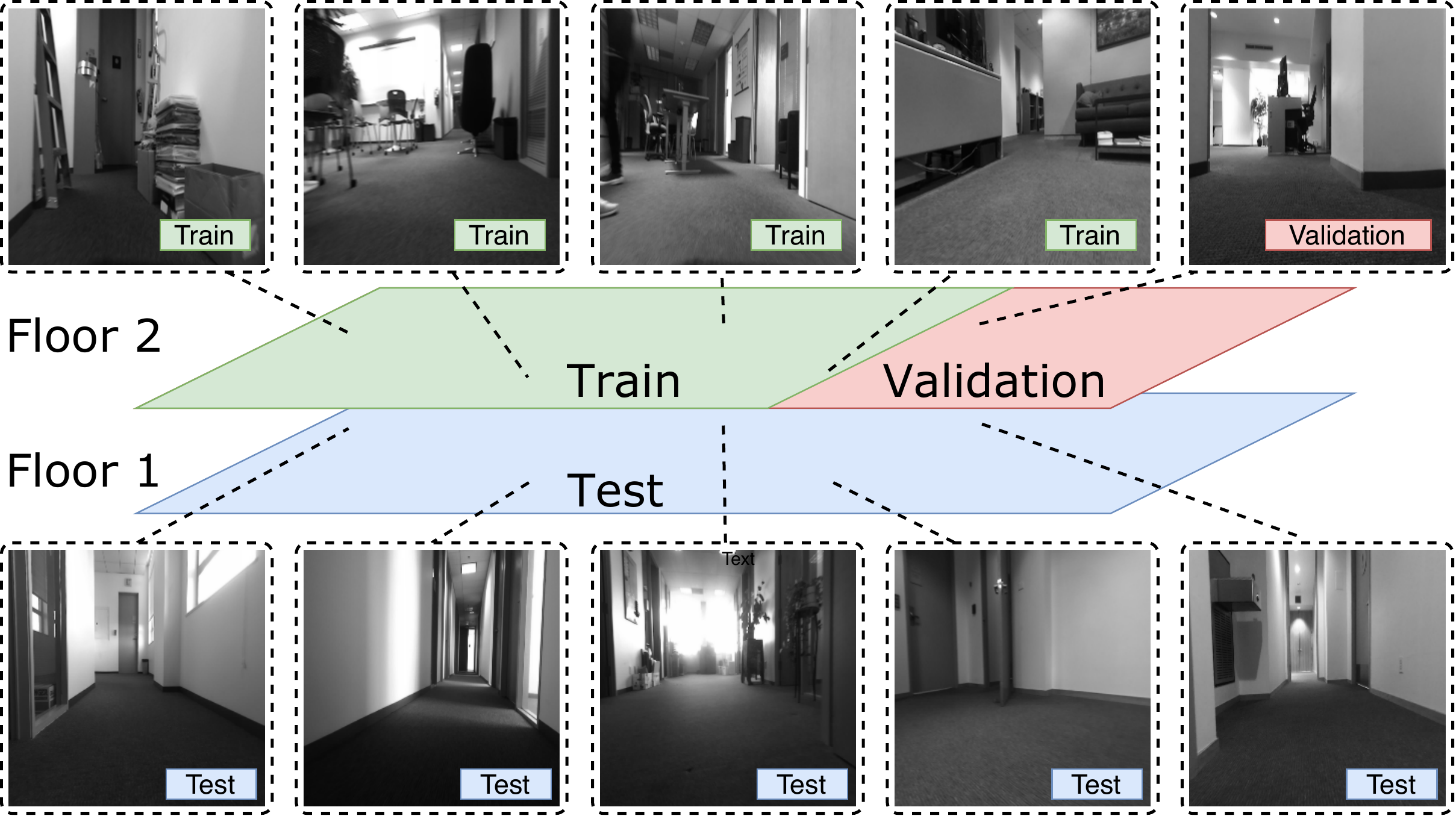}
    \caption{\textit{Overview of our dataset by \cite{christensen2019batvision}.} Training and validation data is collected in separate areas of the same floor, whereas the test data comes from another floor and has different obstacles and decorations.}
    \label{fig:dataset_comparison}
\end{figure}

We use the same dataset as in \cite{christensen2019batvision}, containing time-synchronized binaural audio, RGB images and depth maps for learning associations between sound and vision.
The data has been collected using off-the-shelf, low-cost hardware fitted to a small model car, as shown in Fig.~\ref{fig:overview}. Training, Validation and Test data was collected at different locations of an indoor office with hallways, conference rooms, offices and open areas. The data and collection locations are shown in Fig.~\ref{fig:dataset_comparison}.
We refer to \cite{christensen2019batvision} for more details on signal generation, data collection, hardware and preparation.

\textbf{GCC Features. }We calculate generalized cross-correlation features for pairs of one input channel (left or right ear) respectively and our chirp source waveform:
\begin{align}
\begin{split}
    G_{{l}}(f) &= \frac{X_{{l}}(f)S(f)^*}{|X_{{l}}(f)S(f)^*|}, \\
    G_{{r}}(f) &= \frac{X_{{r}}(f)S(f)^*}{|X_{{r}}(f)S(f)^*|},
\end{split}
\label{eqn:gcc}
\end{align}
where $X_{{l}}(f)$ and $X_{{r}}(f)$ are our left and right waveform represented in the frequency domain, $S(f)$ is our chirp source (described in~\cite{christensen2019batvision}) in the frequency domain padded to the same length as $X(f)$ ($^*$ denotes the complex conjugate). Transformations between the original time-domain and the frequency-domain are obtained by applying the Fourier transformation. The time-domain generalized cross-correlation values are then obtained by applying the inverse Fourier transformation on $G_{{l}}(f)$ and $G_{{r}}(f)$. This is currently a pre-processing step carried out using the {\tt gccphat} tool in MATLAB. In Fig.~\ref{fig:gcc} we show a paired raw waveform sample and its corresponding GCC feature values. The time-series cross-correlation values are then fed to the network, concatenated along the channel dimension.

\section{Proposed Method}
\begin{figure}
    \centering
    \includegraphics[width=1.0\linewidth]{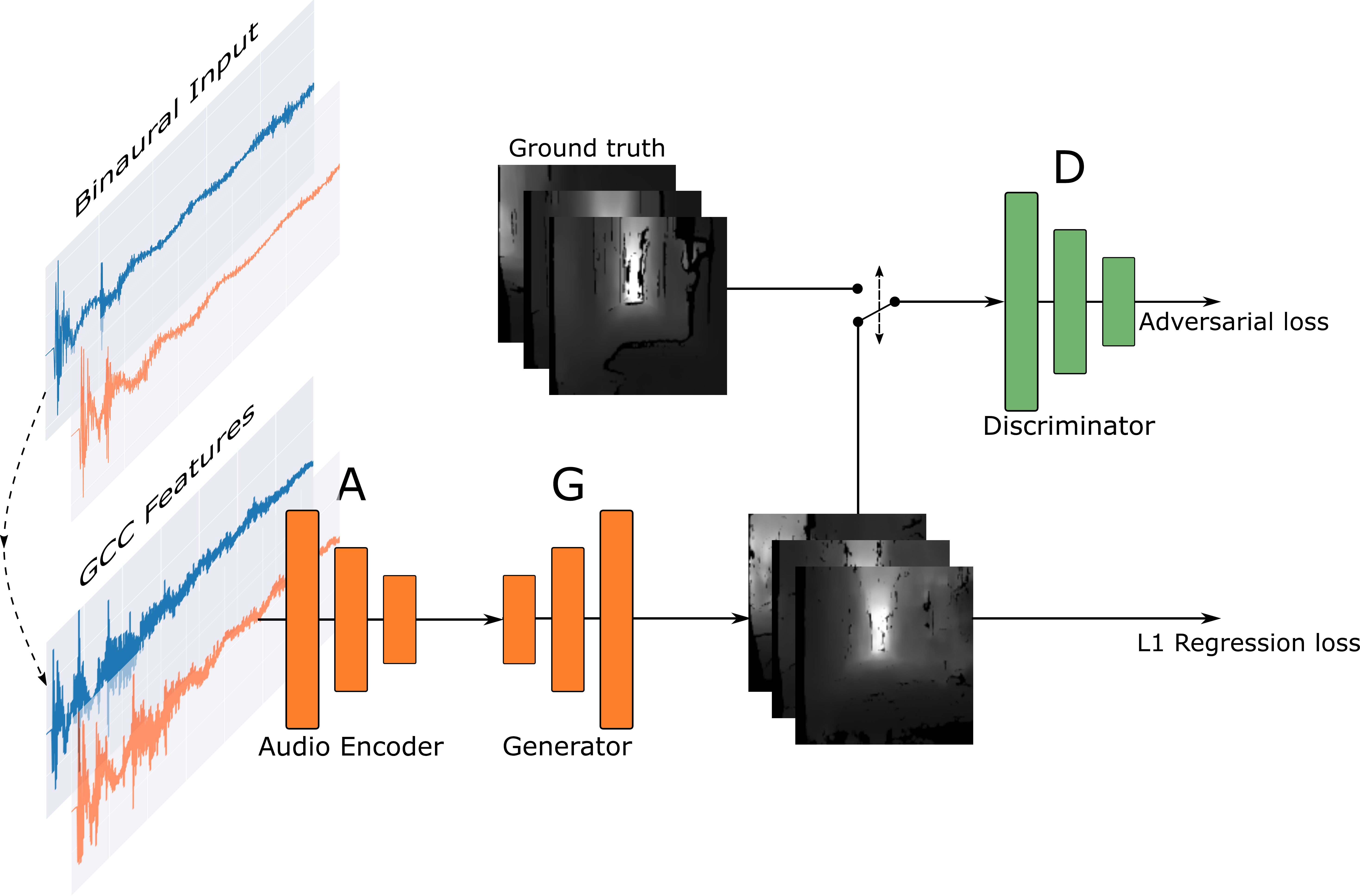}
    \caption{\textit{Our sound to vision network architecture.}  The temporal convolutional audio encoder $A$ turns the binaural input into a latent audio feature vector, based on which the visual generator $G$ predicts the scene depth map. The discriminator $D$ compares the prediction with the ground-truth and enforces high-frequency structure reconstruction at the patch level.
    \label{fig:model}}
\end{figure}

\textbf{Network Architecture. } As shown in our network architecture overview in Fig.~\ref{fig:model}, we keep the high-level design of BatVision. We suggest the following modifications for improving the model and obtaining a more stable training process. 
First, we modify the input to the audio encoder to generalized cross-correlation features rather than raw waveforms or spectrograms of binaural audio signals. 
Second, we re-model the generator and base it on residual learning using Residual-in-Residual Dense Blocks~\cite{DBLP:journals/corr/abs-1809-00219}. 
Third, we replace batch normalization in the discriminator with spectral normalization~\cite{DBLP:journals/corr/abs-1802-05957} and propose a suitable weight factor $\lambda$ for the adversarial loss. 
With these modifications, we observe improved reconstruction results with less artifacts and a more stable training process than in the original model. Please see ~\cite{christensen2019batvision} for details on the original architecture.

The full learning objective of our model is:
\begin{align}
    \min_G \max_D\quad \mathcal{L}_{GAN}(D)+\lambda \mathcal{L}_{GAN}(G)+ \mathcal{L}_{L_1}(G).
\end{align}
$\mathcal{L}_{GAN}$ is a least-squares adversarial loss, $\mathcal{L}_{L_1}$ a $L_1$ regression loss and $\lambda$ a weight factor.

\textbf{Evaluation Metrics. }
For evaluating our predicted depth maps we use a common evaluation method for depth measurements as proposed in~\cite{NIPS2014_5539}. It consist of five evaluation indicators:
\begin{itemize}
    \item Abs Rel $= \frac{1}{\left|N\right|}\sum_{i\in N}\frac{|d_i-d_i^*|}{d_i^*}$,
    \item Sq Rel $= \frac{1}{\left|N\right|}\sum_{i\in N}\frac{||d_i-d_i^*||^2}{d_i^*}$,
    \item RMSE $= \sqrt{\frac{1}{\left|N\right|}\sum_{i\in N}||d_i-d_i^*||^2}$,
    \item RMSE Log $= \sqrt{\frac{1}{\left|N\right|}\sum_{i\in N}||\log(d_i)-\log(d_i^*)||^2}$,
    \item Accuracies: $\textrm{\% of }d_i \, \textrm{ s.t.}\hspace{1em} \textrm{max}\left(\frac{d_i}{d_i^*},\frac{d_i^*}{d_i}\right) = \delta < thr$,
\end{itemize}
where $|N|$ is the total number of pixel with real-depth values, $d_i$ is the predicted depth value of pixel $i$ and $d_i^*$ is the ground truth depth value. Finally, $thr$ denotes a threshold. 

During training we check generated samples constantly in addition to the $L_1$ regression loss of our generator because better perceptual quality of the predictions does not always correspond to lower $L_1$. This is especially true when training in the GAN framework.

\begin{table*}
\centering
\caption{Depth results on our test dataset.}
\label{tab:results}
\begin{tabular}{@{}lccccccc@{}}
\toprule
                               & \multicolumn{4}{c}{Lower is better}                     & \multicolumn{3}{c}{Accuracy: higher is better}            \\ \midrule
                               & Abs Rel & Sq Rel & RMSE & \multicolumn{1}{c|}{RMSE Log} & $\delta < 1.25^1$ & $\delta < 1.25^2$ & $\delta < 1.25^3$ \\ \midrule
\multicolumn{1}{l|}{BatVision~\cite{christensen2019batvision} + Waveforms} & 1.670   & 0.488 &  0.259    & \multicolumn{1}{c|}{\textbf{3.118} }         &      0.249             &    0.359               &   0.484                \\
\multicolumn{1}{l|}{BatVision~\cite{christensen2019batvision} + Spectrograms} & 1.544   & \textbf{0.398} & 0.241    & \multicolumn{1}{c|}{3.177 }         &     0.256             &    0.369               &    0.521                \\
\multicolumn{1}{l|}{BatVision~\cite{christensen2019batvision} + GCC} & 1.782   & 0.464 & 0.252    & \multicolumn{1}{c|}{3.231 }         &      0.236             &    0.330               &    0.454                \\
\multicolumn{1}{l|}{Ours + Waveforms} & 1.839   & 0.472 & 0.245    & \multicolumn{1}{c|}{3.253 }         &      0.252             &    0.357               &    0.471                \\
\multicolumn{1}{l|}{Ours + GCC}      & \textbf{1.542}  &  {0.454}      &  \textbf{0.235}    &  \multicolumn{1}{c|}{3.168}         & \textbf{0.290}                  &   \textbf{0.424}                &     \textbf{0.556}              \\ \bottomrule
\end{tabular}
\end{table*}

\section{Experiments}
We perform all experiments using the same model and hyper-parameters for depth map and grayscale prediction. Also, while training with waveforms or GCC-features we apply the same two input augmentations. 1) We select a window of constant size from the input which start position is randomized in time by 30\%. The center start position is chosen so that always one complete chirp and its echos is captured by the window. We follow the design choices on the length of the window as in \cite{christensen2019batvision} 2) We add Gaussian noise $X\sim \mathcal{N}\left(\mu=0,\sigma^2=\left[0,0.1\right)\right)$ to the signals. The ground truth and predicted output of the model have a spatial size of $128\times 128$.
For our generator, we use a total of 8 Residual-in-Residual Blocks in the low-resolution domain. We follow with a set of up-sampling and convolutional layers until the output resolution is reached. Up-sampling of feature maps is by nearest-neighbor interpolation. We have chosen a batch-size of 16 per GPU, a weight factor $\lambda$ of $1\times 10^{-1}$ and a learning rate of $1\times 10^{-4}$ for both the generator and discriminator. For optimization, we use Adam~\cite{adam_optimizer} with parameters $\beta_1 = 0.5$ and $\beta_2= 0.999$. We alternately update the discriminator and generator until the model generates accurate and visually pleasing results (approx. 100 iterations). We implement our model in the PyTorch framework and train using NVIDIA RTX 2080 TI GPUs.

For a complete comparison we also evaluate the original BatVision network architecture with our proposed GCC-features and our proposed architecture with raw waveforms as input. Using the depth evaluation metrics, we compare depth prediction on the test set of all GAN model combinations and show the results in Table~\ref{tab:results}. A comparison of all trained models and input types is given in Table~\ref{tab:l1}. 

\begin{figure}[h]
    \centering
    \includegraphics[width=1.0\linewidth]{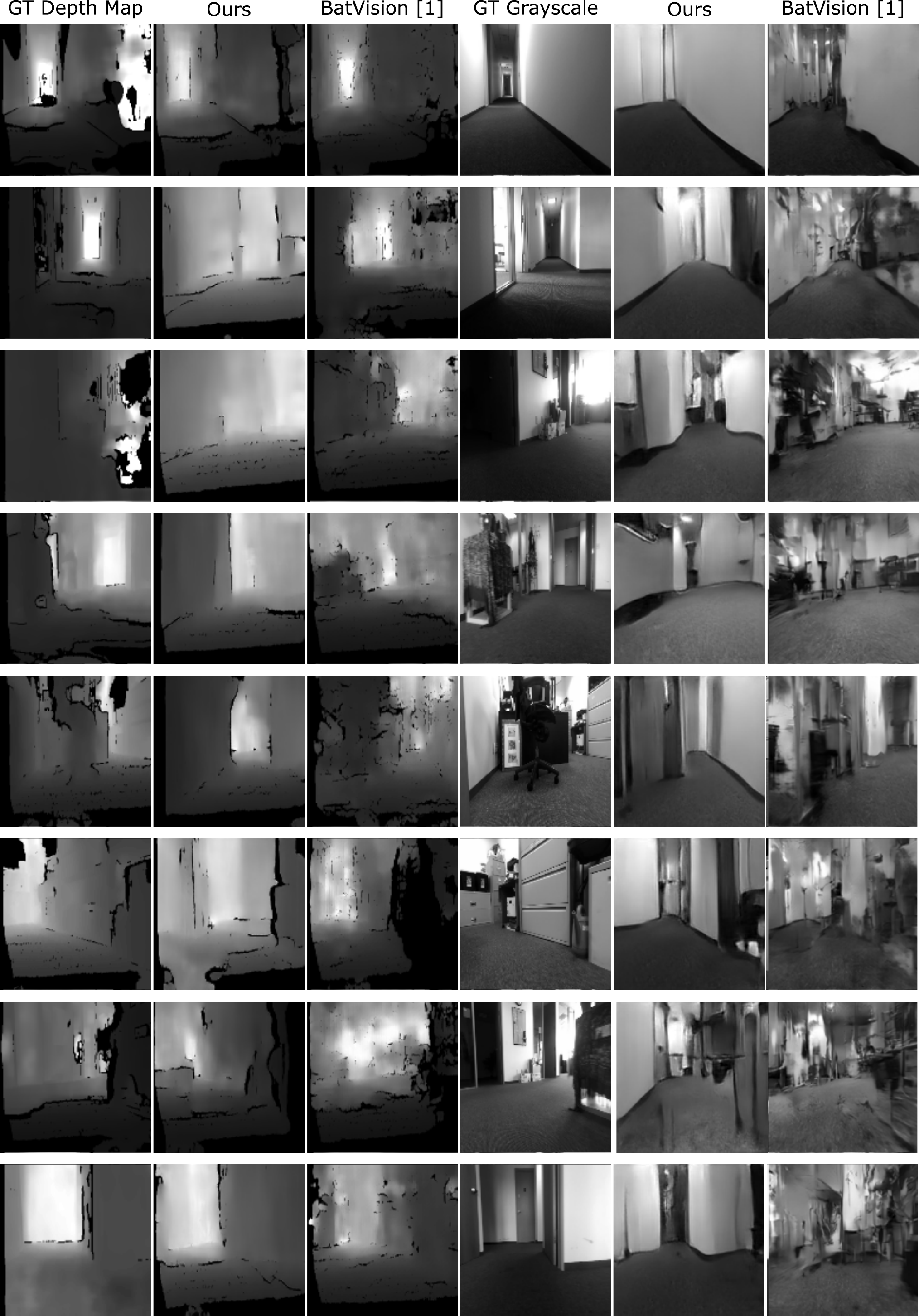}
    \caption{\textit{Test sample reconstructions.}  Columns 1 and 4 show the ground truth depth map and grayscale scene image. The remaining columns show predictions from result from our method and BatVision~\cite{christensen2019batvision}. Overall, our generations show correct mapping of close and distant areas with less noise and more smooth reconstruction than the BatVision method.
    \label{fig:results}}
\end{figure}

We observe that our model improves the original work of BatVision in nearly all metrics and generates qualitatively more accurate and less noisy predictions. In Fig.~\ref{fig:results}, we present examples generated by our best model as indicated by Table~\ref{tab:l1}~and~\ref{tab:results}. Note that it is not possible to predict the exact grayscale image of a scene because not all information about appearance can be transported by sound. Rather the goal is to reconstruct an image which shows plausible layout in terms of free space and obstacles. Finally, when training both approaches we empirically find that our proposed method is more stable during training and less affected by small changes in hyper-parameters compared to BatVision.

\begin{table}
\centering
\caption{$L_1$ loss on the test set for depth map and grayscale generation for different network configurations.}
\label{tab:l1}
\begin{tabular}{@{}lcc@{}}
\toprule
\multicolumn{1}{l|}{\textit{Arch. + Input}} & \multicolumn{2}{c}{$L_1$ Loss}          \\ \midrule
\multicolumn{1}{c|}{\textit{Depth Map}}     & \multicolumn{1}{c|}{\textit{Gen. Only}} & \textit{GAN} \\ \midrule
\multicolumn{1}{l|}{BatVision~\cite{christensen2019batvision} + Waveforms}        & \multicolumn{1}{c|}{0.0880}          &   0.0930  \\
\multicolumn{1}{l|}{BatVision~\cite{christensen2019batvision} + Spectrograms}        & \multicolumn{1}{c|}{0.0742}          &    0.0878  \\
\multicolumn{1}{l|}{BatVision~\cite{christensen2019batvision} + GCC}  & \multicolumn{1}{c|}{0.0678}          & 0.0758     \\
\multicolumn{1}{l|}{Ours + Waveforms} & \multicolumn{1}{c|}{0.0698}          &   0.0773  \\
\multicolumn{1}{l|}{Ours + GCC}       & \multicolumn{1}{c|}{\textbf{0.0645}}          &  \textbf{0.0732}   \\ \midrule
\multicolumn{1}{c|}{\textit{Grayscale}} & \multicolumn{2}{c}{\textit{GAN}} \\ \midrule
\multicolumn{1}{l|}{BatVision~\cite{christensen2019batvision} + Waveforms} & \multicolumn{2}{c}{0.2018} \\
\multicolumn{1}{l|}{BatVision~\cite{christensen2019batvision} + Spectrograms} & \multicolumn{2}{c}{0.1841} \\
\multicolumn{1}{l|}{Ours + GCC} & \multicolumn{2}{c}{\textbf{0.1770}} \\
\bottomrule
\end{tabular}
\end{table}


\section{Conclusions}
We evaluated generalized cross-correlations features over raw waveforms as input modality and novel model configurations for BatVision \cite{christensen2019batvision}. With \textit{Residual-in-Residual} Dense Blocks in the generator and spectral normalization in the discriminator we achieve major quantitative and qualitative improvements. Apart from better scores on the evaluation metric, reconstructed depth and grayscale images show significantly better perceptual quality. The results in this work show as proof-of-concept the potential information, contained in sound. Complementary to vision we argue it can be useful in many tasks, either as exclusive or additional sensor input or to guide machine learning, as recently well presented in concurrent work \cite{gao2020visualechoes}.

{\small
\bibliographystyle{ieee_fullname}
\bibliography{egbib}
}

\end{document}